%% file: 0358.tex
\documentclass[runningheads]{llncs}
\usepackage{graphicx}
\usepackage{amsmath,amssymb} 
\usepackage{color}

\usepackage{hyperref}
\usepackage[dvipsnames]{xcolor}
\usepackage{multirow}
\usepackage[font={small}]{caption}
\usepackage{subcaption}
\usepackage{tablefootnote}
\usepackage[symbol]{footmisc}
\usepackage{booktabs}
\usepackage{float}

\def\eg{\emph{e.g.}} 
\def\ie{\emph{i.e.}} 
\def\etal{\emph{et al.}}
\def\wrt{w.r.t.}

\begin{document}
\title{Diagnosing Error in Temporal Action Detectors}

\titlerunning{Diagnosing Error in Temporal Action Detectors}
%
\author{Humam Alwassel \and
Fabian Caba Heilbron \and
Victor Escorcia \and 
Bernard Ghanem}
%
\authorrunning{H. Alwassel \etal}
%

\institute{King Abdullah University of Science and Technology (KAUST), Saudi Arabia\\
\url{http://www.humamalwassel.com/publication/detad/} \\
\email{\{humam.alwassel, fabian.caba, victor.escorcia, bernard.ghanem\}@kaust.edu.sa}
}
\maketitle
%
\begin{abstract}
    Despite the recent progress in video understanding and the continuous rate of improvement in temporal action localization throughout the years, it is still unclear how far (or close?) we are to solving the problem.
    To this end, we introduce a new diagnostic tool to analyze the performance of temporal action detectors in videos and compare different methods beyond a single scalar metric.
    We exemplify the use of our tool by analyzing the performance of the top rewarded entries in the latest ActivityNet action localization challenge.
    Our analysis shows that the most impactful areas to work on are: strategies to better handle temporal context around the instances, improving the robustness \wrt~the instance absolute and relative size, and strategies to reduce the localization errors.
    Moreover, our experimental analysis finds the lack of agreement among annotator is not a major roadblock to attain progress in the field.
    Our diagnostic tool is publicly available to keep fueling the minds of other researchers with additional insights about their algorithms.{\let\thefootnote\relax\footnote{{The first three authors contributed equally to this work. Authors ordering was determined using Python's random.shuffle() seeded with the authors' birthday dates.}}}
\end{abstract}

\input{sections/intro.tex}

\input{sections/preliminaries.tex}

\input{sections/user_study.tex}

\input{sections/taxonomization.tex}

\input{sections/fp_analysis.tex}

\input{sections/sensitivity.tex}

\input{sections/fn_analysis.tex}

\input{sections/discussion.tex}

\clearpage

\bibliographystyle{splncs04}
\bibliography{mybib}

\end{document}

%% file: sections/intro.tex
\section{Introduction}
We are in the \textit{Renaissance} period of video understanding.
Encouraged by the advances in the image domain through representation learning \cite{GirshickDDM16,HeZRS16,NIPS2012_4824}, large scale datasets have emerged over the last couple of years to challenge existing ideas and enrich our understanding of visual streams \cite{GoyalKMMWKHFYMH17,GuSVPRTLRSSM17,HeilbronEGN15,IdreesZJGLSS17,KarpathyTSLSF_cvpr14,KayCSZHVVGBNSZ_2017,monfortmoments,SigurdssonVWFLG16}.
Recent work has already shown novel algorithms \cite{WangGGH_2017} and disproved misconceptions associated with underrated 3D representations for video data \cite{CarreiraZ17}.
However, we are still awaiting the breakthrough that allows us to temporally localize the occurrence of actions in long untrimmed videos \cite{GhanemNSCAKEHB_2017,IdreesZJGLSS17}.
In this paper, we propose to step back and analyze the recent progress on the temporal action localization as a means to fuel the next generation with the right directions to pursue.

\begin{figure}[t!]
    \centering
    \includegraphics[width=0.99\textwidth]{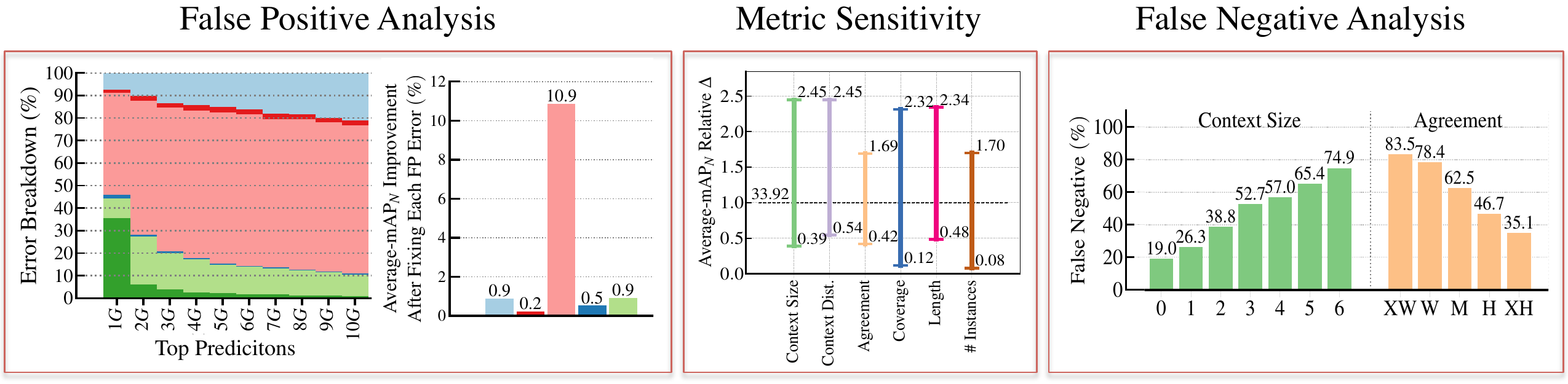}
    \caption{Illustration of the three types of analyses that our diagnostic tool provides for action localization algorithms. \textbf{Left:} We analyze the false positive error sources and their impact on the performance. \textbf{Middle:} We investigate the localization metric sensitivity to different characteristics of the ground truth instances. \textbf{Right:} We inspect the influence of ground truth instance characteristics on the miss detection rate.}
    \label{fig:pull_figure}
\end{figure}

Currently, researchers have appealing intuitions to tackle video action localization problem \cite{AlwasselHG18,EscorciaDJGS18,scnn,ZhaoXWWTL_iccv2017}, they are equipped with large datasets to validate their hypothesis \cite{HeilbronEGN15,IdreesZJGLSS17,HeilbronJHG18}, and they have access to appropriate computational power. 
Undoubtedly, these aspects helped materialize an increasing performance throughout the years \cite{scc,GhanemNSCAKEHB_2017,ZhaoXWWTL_iccv2017}. Yet, such improvements are not enough to describe the whole picture. For example, we are still not able to answer the following questions: How close are we to achieve our goal of delimiting the start and end of actions? What makes an algorithm more effective than another? What makes an action hard to localize? Is the uncertainty of the temporal boundaries impeding the development of new algorithms? Inspired by similar studies in other areas \cite{HoiemCD12,RonchiP17,SigurdssonRG17,ZhangBOHS16}, we take a deep look at the problem beyond a single scalar metric and perform a quantitative analysis that: (i) informs us about the kind of errors a given algorithm makes, and measures the impact of fixing them; (ii) describes which action characteristics impact the performance of a given algorithm the most; and (iii) gives insights into the action characteristics a proposed solution struggles to retrieve. Figure \ref{fig:pull_figure} shows a brief glimpse of our diagnostic analysis applied to a state-of-the-art method on ActivityNet version 1.3 \cite{HeilbronEGN15,LinZS_2017}.

\noindent \textbf{Relation to Existing Studies.}
The seminal work of Hoiem \etal~showcased the relevance of diagnosing the failure modes of object detectors in still images \cite{HoiemCD12}.
Inspired by this work, \cite{EveringhamEGWWZ15,IdreesZJGLSS17,MoltisantiWMD17,RonchiP17,RussakovskyDHBF13,RussakovskyDSKS15,SigurdssonRG17,ZhangBOHS16} provided insightful analysis of algorithms for multiple localization tasks such as human pose estimation, object detection at large-scale and multi-label action classification, in still images in most of the cases.
In contrast with them, our work contributes to the understanding of temporal action localization in untrimmed videos.

\noindent \textit{Object detection and human pose estimation.}
\cite{HoiemCD12} pioneered the categorization of localization errors as a mean to get more insights about the performance of object detection algorithms. \cite{RonchiP17,ZhangBOHS16} extended the diagnostic tools to the context of human pose estimation showing the relevance of this approach to quantitatively identify the failure modes and to recommend ways to improve existing algorithms for body parts localization.
In a similar spirit, our work is the first that characterizes the localization errors for temporal action localization in videos.

\noindent \textit{Multi-label action classification.}
Sigurdsson \etal~\cite{SigurdssonRG17} provides an insightful diagnosis of algorithms and relevant directions needed for understanding actions in videos.
\cite{SigurdssonRG17} studies the influence of different attributes, such as verbs, objects, human poses, and the interactions between actions in the scope of video action recognition. Most of the study is centered around action classification at the frame level or the entire video and is carried on relatively short streams of $30$ seconds on average.
Our work contributes with an orthogonal perspective to this study, performing an in-depth analysis of the problem of delimiting temporal boundaries for actions in long videos.

\noindent\textbf{Contributions.} 
Our contributions in this paper are threefold.
\textbf{(i)} We collect additional annotation data for action context and temporal agreement in ActivityNet. We use the collected data to categorize the ground truth instances into six action characteristics: context size, context distance, agreement, coverage, length, and the number of instances (Section \ref{section:dataset_characterization}).
\textbf{(ii)} We investigate and classify the most relevant error types to temporal action localization (Section \ref{section:taxonomization}).
\textbf{(iii)} We provide a complete analysis tool (annotations, software, and techniques) that facilitates detailed and insightful investigation of action detectors performance. We exemplify the use and capabilities of our diagnosis tool on the top four action detectors in the recent ActivityNet 2017 challenge (Sections \ref{section:fp_analysis} -\ref{section:fn_analysis}).

%% file: sections/preliminaries.tex
\section{Preliminaries}

\noindent\textbf{Evaluation Framework.}
We use the ActivityNet dataset v1.3 \cite{HeilbronEGN15} as a test bed for our diagnostic analysis of the progress in temporal action localization in videos.
The choice of this dataset obeys multiple reasons,
(i) it is a large scale dataset of $20$K videos with an average length of four minutes;
(ii) it consists of a diverse set of human actions ranging from household activities, such as \textit{washing dishes}, to sports activities, like \textit{beach volleyball}. This allow us to make conclusions about a diverse type of actions;
(iii) it is an active non-saturated benchmark with a held-out test set and an additional validation set, ensuring good machine learning practices and limiting over-fitting risk;
(iv) it provides an open-source evaluation framework and runs an annual competition, which safeguards good progress on the community.
Additionally, we extend our analysis to the widely used THUMOS14 dataset \cite{THUMOS14} in the \textit{supplementary material}.
In this way, we cover the most relevant benchmarks used to dictate the progress in this area. 

The action localization problem measures the trade-off that an algorithm consistently retrieves the occurrence of true action instances, from different classes, without increasing the numbers of spurious predictions. This task is evaluated by measuring the precision and recall of the algorithms.
The metric used to trade-off precision and recall for retrieving the segments of a particular action is the Average Precision (AP), which corresponds to an interpolated area under the precision-recall curve \cite{EveringhamEGWWZ15}. 
To evaluate the contribution of multiple action classes, the AP is computed independently for each category and averaged to form the mean AP (mAP).
Given the continuous nature of the problem, a prediction segment is considered a true positive if its temporal Intersection over Union (tIoU) with a ground truth segment meets a given threshold. To account for the varied diversity of action duration, the public evaluation framework employs the average-mAP, which is the mean of all mAP values computed with tIoU thresholds between $0.5$ and $0.95$ (inclusive) with a step size of $0.05$.

To establish a middle ground between multiple algorithms that is robust to variations of ratio between true and false positives across multiple classes, we employ the normalized mean AP \cite{HoiemCD12}. In this way, we can compare the average-mAP between uneven subsets of ground truth instances, \eg~when the number of instances of a given category doubles the number of instances of another category for a given detection rate. We compute the normalized mAP (mAP$_{N}$) in terms of the normalized precision $P_N(c) = \frac{R(c)\cdot N}{R(c) \cdot N + F(c)}$,
where $c$ is the confidence level, $R(c)$ is the recall of positive samples with confidence at least $c$, $F(c)$ is the false positive rate for predictions with confidence at least $c$, and $N$ is a constant number. We report average-mAP$_N$ as the the action localization metric, and set $N$ to the average number of ground truth segments per class.

\begin{table}[t!]
    \begin{center}
        \tabcolsep=0.25cm
        \caption{
            Localization performance as measured by average-mAP and average-mAP$_{N}$ on ActivityNet \cite{HeilbronEGN15}. We show the two metrics for all predictions and for the top-$10G$ predictions, where $G$ is the number of ground truth instances. Using average-mAP$_{N}$ gives slightly higher values. Notably, limiting the number of predictions to the top-$10G$ gives performance values similar to those when considering all predictions.
        }
        \label{table:ANET_mAP_normalized}
        \fontsize{7}{8.2}\selectfont
        \begin{tabular}{ l || c c | c c}
            \toprule
                            & \multicolumn{2}{c|}{\textbf{Average-mAP} (\%)} & \multicolumn{2}{c}{\textbf{Average-mAP$_N$}(\%)} \\
            \textbf{Method} & All & top-$10G$ & All & top-$10G$\\
            
            \hline
            \hline
            
            SC  & $33.42$          & $32.99$          & $33.92$          & $33.45$ \\
            CES & $31.87$          & $31.83$          & $32.24$          & $32.20$ \\
            IC  & $31.84$          & $31.70$          & $32.14$          & $32.00$ \\
            BU  & $16.75$          & $16.52$          & $17.26$          & $17.02$ \\
            \bottomrule
            
        \end{tabular}
    \end{center}    
\end{table}

\noindent\textbf{Algorithms.}
We exemplify the use of our diagnostic tool by studying the four rewarded approaches in the latest action localization task in the ActivityNet challenge \cite{GhanemNSCAKEHB_2017} (Table \ref{table:ANET_mAP_normalized} summarizes the methods' performances).
Interestingly, all the methods tackled the problem in a two-stage fashion, using a proposal method \cite{BuchESGN17,EscorciaHNG16,GaoYSCN_iccv2017,HeilbronNG16,scnn} followed by a classification scheme \cite{SimonyanZ_nips2014,TranBFTP_iccv15,WangXWQLTG_eccv2016}. However, there are subtle design differences which are relevant to highlight.

\textit{SC} \cite{LinZS_2017}. It was the winner of the latest action localization challenge with a margin of $2\%$ average-mAP. The key ingredient for its success relies on improving the action proposals stage. To this end, this work re-formulates the fully convolutional action detection network SSAD \cite{LinZS_acm2017} as a class-agnostic detector.
The detector generates a dense grid of segments with multiple durations, but only those near the occurrence of an instance receive a high score.
In addition to the multi-scale proposal network, this work refines proposals' boundaries based on the outputs of the TAG grouping approach \cite{XiongZWLT_2017}.
Finally, the classification stage is performed at the video level independently of the proposal stage results.

\textit{CES} \cite{GhanemNSCAKEHB_2017,ZhaoXWWTL_iccv2017}. This work achieved the runner-up entry in the challenge \cite{GhanemNSCAKEHB_2017}, and held the state-of-the-art approach on THUMOS14 at that time. It employs a temporal grouping heuristic for generating actions proposals \cite{XiongZWLT_2017} from dense actioness predictions. The proposals are classified and refined in a subsequent stage by the SSN network \cite{ZhaoXWWTL_iccv2017}. Most of its effort involves enhancing the SSN network to a diverse set of actions.
SSN applies a temporal pyramid pooling around the region spanned by a proposal segment, and then it classifies the segment by balancing the information inside the segment and the context information around it. This work found consistent improvement in the validation set through the use of deeper architecture and fine tuning on the larger Kinetics dataset \cite{KayCSZHVVGBNSZ_2017}.

\textit{IC} \cite{GhanemNSCAKEHB_2017}. This approach ranks third by employing a similar strategy to the CES submission. Its main distinction relies on using a sliding window-based proposal scheme as well as employing human pose estimation to influence the classification decisions of the SSN network \cite{ZhaoXWWTL_iccv2017}.

\textit{BU} \cite{XuD_iccv2017}. It was awarded the challenge most innovative solution. 
This work extended the Faster RCNN architecture \cite{RenHGS_nips2015} to the problem of temporal action localization. It designed a temporal proposal network coupled with a multi-layer fully connected network for action classification and boundary refinement.
In comparison with the top-ranked submissions that exploit optical flow or human pose estimation, this work relies solely on RGB streams to learn a temporal representation via 3D convolutions pretrained on Sports-1M dataset \cite{KarpathyTSLSF_cvpr14}.

%% file: sections/user_study.tex
\begin{figure}[t!]
    \begin{minipage}{0.49\textwidth}
        \centering
        \includegraphics[width=1\linewidth]{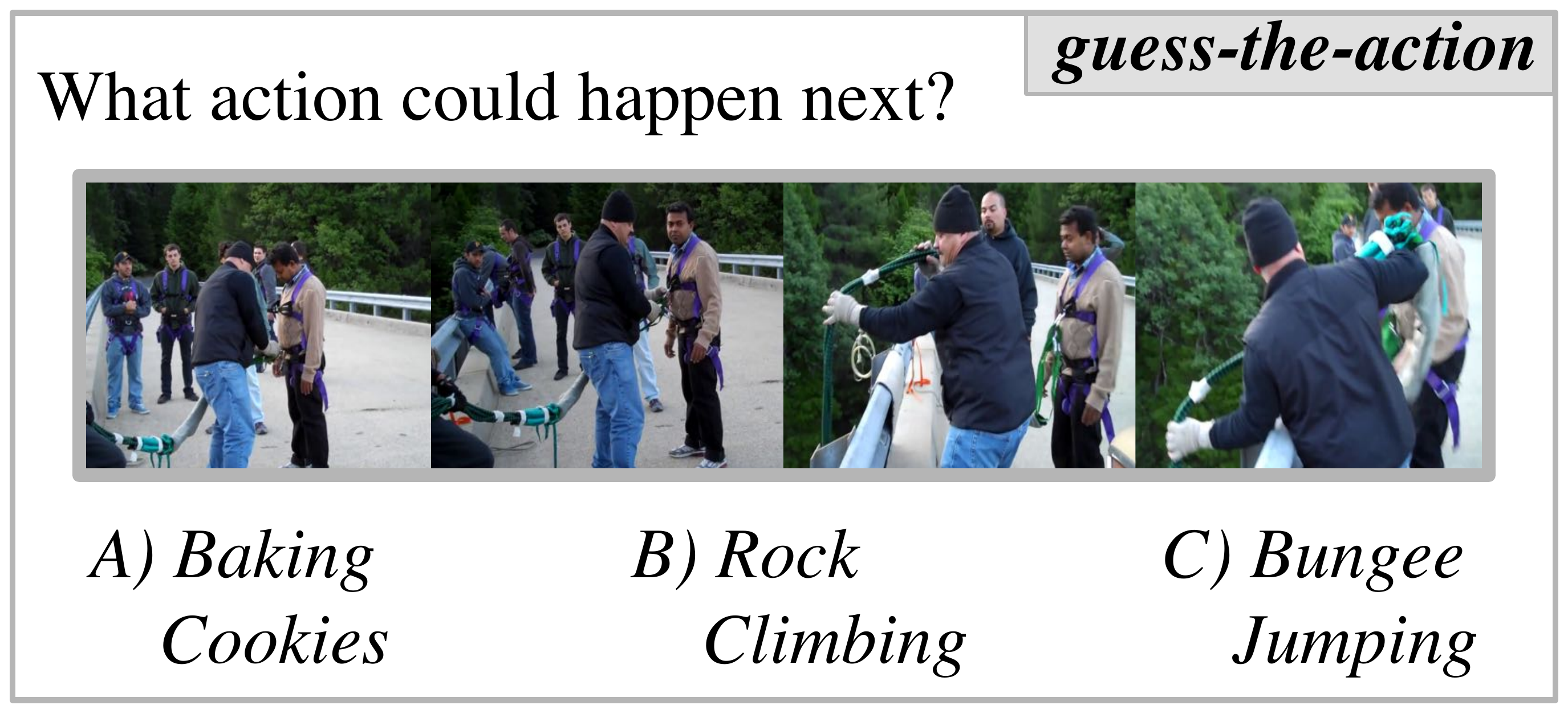}
    \end{minipage}\hfill
    \begin{minipage}{0.49\textwidth}
        \centering
        \includegraphics[width=1\linewidth]{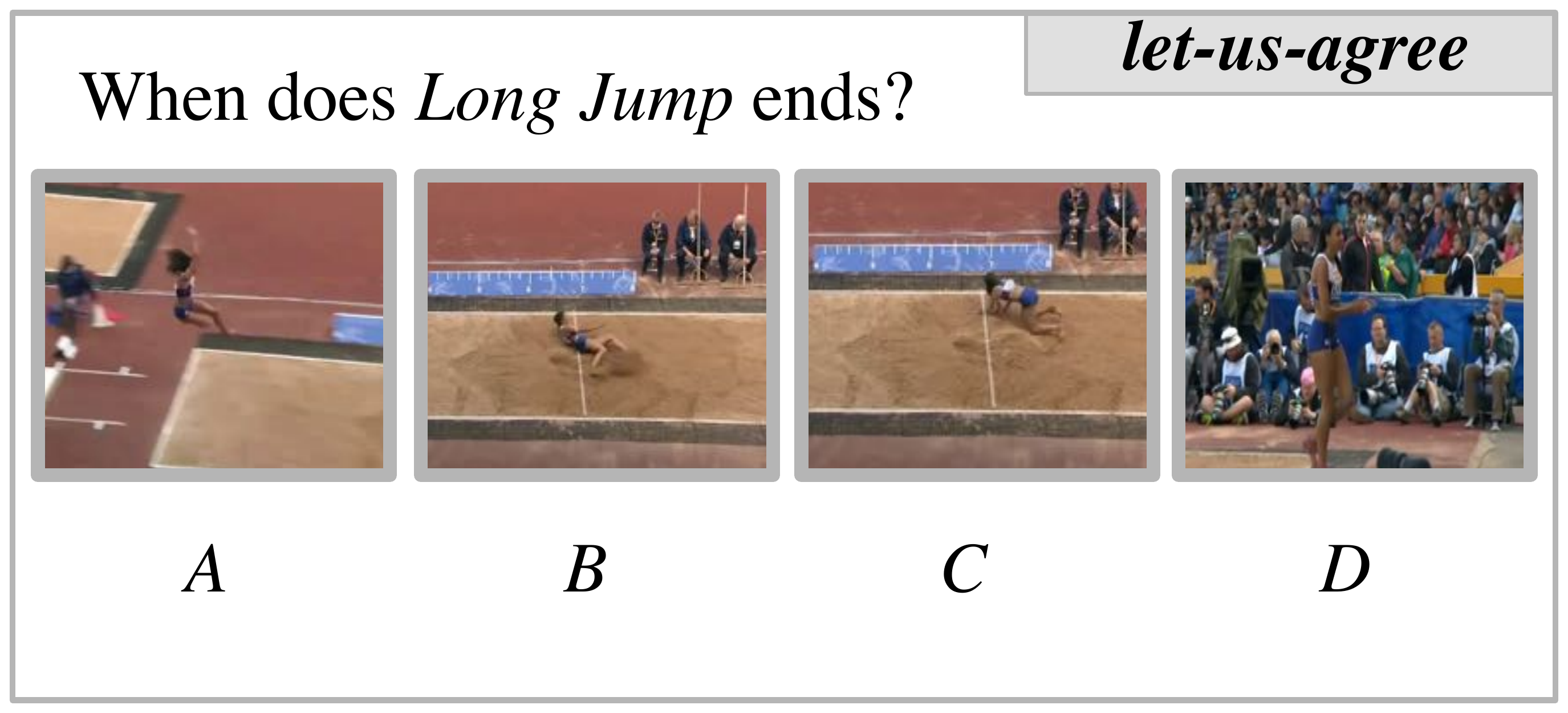}
    \end{minipage}
    \caption[]{\textbf{Left:} \textit{guess-the-action} game. In this game you have to guess what action (one out of three options) could happen in the context of the depicted video clip. \textbf{Right:} \textit{let-us-agree} game. Here, the goal is to pick the frame that best represents when the action \textit{Long Jump} ends. To check your answers read the footnote\footref{footnote:game_answer}.}
    \label{fig:games}
\end{figure}

\section{Dataset Characterization} \label{section:dataset_characterization}

Our first goal is to describe datasets with inherent characteristics such as coverage, length, and the number of instances. Moreover, we are interested in augmenting the dataset with two additional characteristics, temporal context and temporal boundary agreement, which we argue are critical for understanding the current status of action localization. Let's play some games to motivate our selection (Jump to Figure \ref{fig:games}). The first game, \textit{guess-the-action}, consists of watching a series of frames to guess what action happens next. The second game, \textit{let-us-agree}, asks you to pick the instant when a given action ends. We invite you to play the game and check your answers afterwards in the footnote\footnote{\label{footnote:game_answer} (1): The action that happens is \textit{Bungee Jumping}. (2): There is not a unanimous answer for this game. $67\%$ of our lab colleagues picked frame B as the correct answer.}. We relate the first game with whether an action instance has temporal context or not. If an action instance is in temporal context, the player should be able to exploit semantic information such as objects, scenes, or motions to guess what action happens either before or after. The second game explores how humans agree on defining temporal boundaries of an action instance. Surprisingly, this toy example reveals that defining an action's temporal boundaries is hard. Intrigued, we decided two conduct two formal online user studies with the aim of quantifying the amount of temporal context and temporal boundaries agreement for temporal action localization. In this section, we first present the online user studies that allow us to augment ActivityNet v1.3 with the temporal context and temporal boundaries agreement attributes. Then, we provide a detailed definition of each action characteristics studied in this work.

\subsection{Online User Studies}
\noindent\textbf{User Study I: Temporal Context of Actions.}
Our goal is to quantify the amount of temporal context around an action instance. To that end, we conduct an online user study that resembles the \textit{guess-the-action} game described earlier. We choose Amazon Mechanical Turk as a test bed to hold the user study. Each participant's task is to watch a $5$-second video clip and pick, from a given list, all the human actions that they believe could happen in the context of the video clip. We revisit our definition of temporal context, which describes that an action instance is in temporal context if semantic information \textit{around} the instance helps a person to guess the action class of such instance. Thus, we investigate the temporal context of an instance by sampling six non-overlapping $5$-second clips around the action's temporal boundaries. 
We present each user with three different candidate classes, one of the options is the correct action class, and the other two options are either similar or dissimilar class to the ground truth class. 
Following the findings of \cite{scc}, we use objects and scene information to form sets of similar and dissimilar actions.
Given that multiple selections are allowed, we consider an answer as correct if the participant chooses the correct action only, or if they pick the correct action and the option that is similar to it. If a temporal segment allows the participant to guess the action, we call that segment a \textit{context glimpse}.

Our study involved $53$ Amazon Mechanical Turk workers (Turkers), who spent a median time of $21$ seconds to complete a single task. In total, we submitted a total of $30$K tasks to cover the existing instances of ActivityNet. Interestingly, Turkers were able to correctly guess the action in $90.8\%$ of the tasks. While that result can be interpreted as a signal of dataset bias towards action-centric videos, it also suggests that action localization methods would require temporal reasoning to provide accurate predictions in such scenario. For instance, most probably you used information about scene (bridge, river) and objects (elastic cord, helmet) to predict the \textit{Bungee Jumping} answering when playing \textit{guess-the-action}. However, such high-level information did not help you to provide the ending time of \textit{Long Jump} in the \textit{let-us-agree} game. In short, for each ActivityNet temporal instance, we conducted $6$ temporal context experiments, which we use later when defining action characteristics.

\noindent\textbf{User Study II: Finding Temporal Boundaries of Actions.} After playing \textit{let-us-agree}, the question naturally arises, can we precisely localize actions in time? To address this question, we followed \cite{SigurdssonRG17} and designed an instance-wise procedure that helped us characterize the level of human agreement achieved after annotating temporal bounds of a given action.

We relied on $168$ Turkers to \textit{re-annotate} temporal boundaries of actions from ActivityNet. The median time to complete the task was three minutes. The task consisted in defining the boundaries of an already spotted action. Additionally, we asked the participants to annotate each temporal boundary \textit{individually}. For each action instance, we collected three new annotations from different Turkers. We measure agreement as the median of the pairwise tIoU between all the four annotations (the original annotation and the three newly collected ones). As a result, Turkers exhibited an agreement score of $64.1\%$ over the whole dataset. The obtained results suggests that it is hard to agree, even for humans, about the temporal boundaries of actions, which matches with previously reported conclusions \cite{SigurdssonRG17}. In summary, we collected three additional annotations for each action instance from ActivityNet, enabling future discussions about the effect of ambiguous boundaries on action detectors.

\subsection{Definitions of Action Characteristics}

\begin{figure}[t!]
    \centering
    \begin{subfigure}{1\linewidth}
        \includegraphics[width=0.99\textwidth]{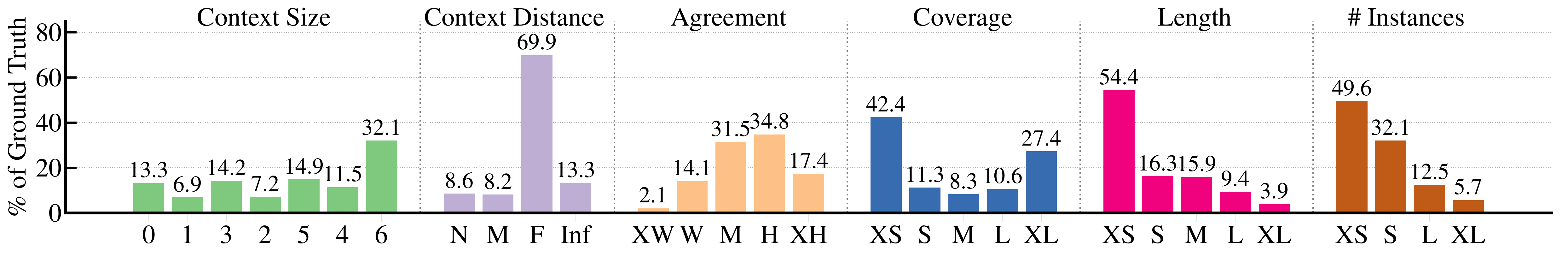}
    \end{subfigure}
    \caption{Distribution of instance per action characteristic. We report the percentage of ground truth instances belonging to each characteristic bucket.}
    \label{fig:dataset_characterization}
\end{figure}

We annotate each instance of the ActivityNet v1.3 dataset with six different characteristics: context size, context distance, agreement, coverage, length, and number of instances. 
Here, we define these characteristic and discuss their distribution (Figure \ref{fig:dataset_characterization}).

\noindent\textbf{Context Size.} We use the collected data from User Study I to characterize the amount of temporal context around an instance. We define context size as the number of \textit{context glimpses} associated with an instance. Thus, values of context size range from $0$ to $6$. Interestingly, we find that only $6.9\%$ of instances do not have temporal context. Additionally, many instances have large temporal context, \eg~$58.4\%$ of instances have more than $3$ context glimpses.

\noindent\textbf{Context Distance.} We use the results from User Study I to characterize the furthest distance away from the instance where a \textit{context glimpse} exists. We define four types of context distance: Inf, which indicates that no temporal context exists; Far (F); Middle (M); Near (N). Notably, We see that most instances ($69.9\%$) have \textit{context glimpses} far away.

\noindent\textbf{Agreement.} Our goal is to characterize an instance based on how difficult it is to agree on its temporal boundaries. To this end, we exploit the data collected from User Study II. We measure agreement as the median tIoU between all annotation pairs for an instance. We form five groups based on agreement score (median tIoU):
Extra Weak (XW: $(0,0.2]$), Weak (W: $(0.2,0.4])$), Mid (M: $(0.4,0.6]$), High (H: $(0.6,0.8]$), and Extra High (XH: $(0.8,1.0]$). We discover that a relatively small number of instances have extremely weak agreement ($2.1\%$). On the other hand, most of the dataset ($83.8\%$ of instances) exhibit at least Mid agreement.

\noindent\textbf{Coverage.} To measure coverage, we normalize the length of the instance by the duration of the video. We categorize coverage values into five buckets: 
Extra Small (XS: $(0,0.2]$), Small (S: $(0.2,0.4]$), Medium (M: $(0.4,0.6]$), Large (L: $(0.6,0.8]$), and Extra Large (XL: $(0.8, 1.0]$). Interestingly, Extra Small and Extra Large instances compose most of the dataset with $42.4\%$ and $27.4\%$ of instances assigned to each bucket, respectively.

\noindent\textbf{Length.} We measure length as the instance duration in seconds. We create five different length groups: 
Extra Small (XS: $(0,30]$), Small (S: $(30,60]$), Medium (M: $(60,120]$), Long (L: $(120,180]$), and Extra Long (XL: $>180$). 
We find that more than half ($54.4\%$) of the instances are small. We also observe that the instance count gradually decrease with length size.

\noindent\textbf{Number of Instances (\# Instances).} We assign each instance the total count of instances (from the same class) in its video. We create four categories for this characteristic: Extra Small (XS: $1$); Small (S: $[2,4]$); Medium (M: $[5,8]$); Large (L: $>8$). We find half of the dataset contains a single instance per video.

%% file: sections/taxonomization.tex
\section{Categorization of Temporal Localization Errors}\label{section:taxonomization}

When designing new methods, researchers in the field often identify an error source current algorithms fail to fully address. For example, \cite{cdc} identifies the problem of localization errors at high tIoU thresholds and devises the CDC network to predict actions at frame-level granularity. However, the field lacks a detailed categorization of the errors of specific relevance to the temporal localization problem. A thorough classification of error types and analysis of their impact on the temporal localization performance would help guide the next generation of localization algorithms to focus on the most significant errors. To this end, we propose in this section a taxonomy of the errors relevant to action localization, and we analyze the impact of these errors in Sections \ref{section:fp_analysis} and \ref{section:fn_analysis}.

\begin{figure}[t!]
    \centering
    \includegraphics[width=\linewidth]{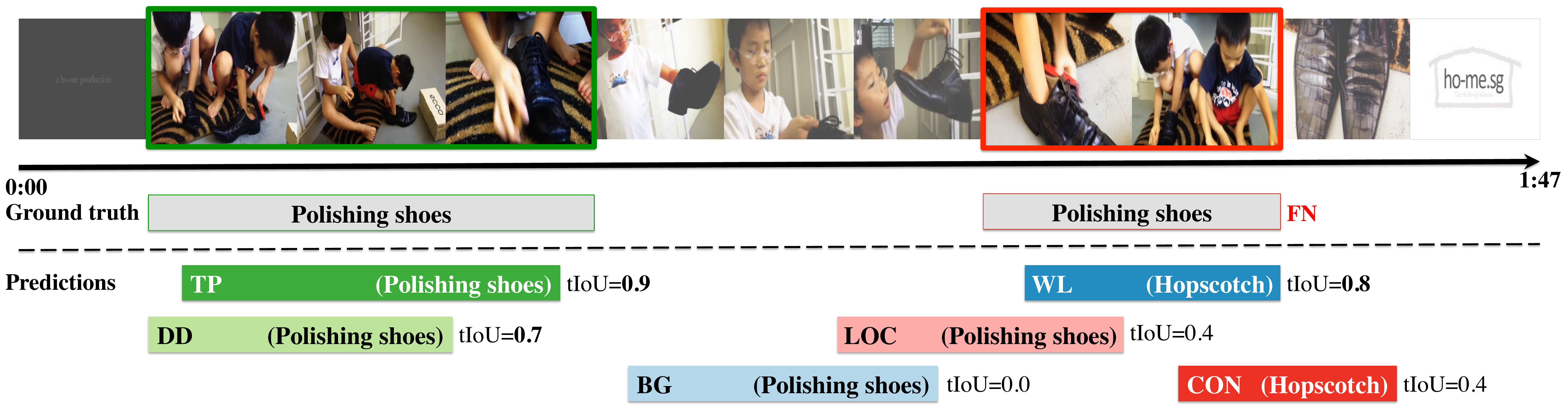}
    \caption{Illustration of the most relevant action localization errors (Section \ref{section:taxonomization}). Predictions with bold tIoU values meet the tIoU threshold ($0.55$ in this example). The left action instance is correctly matched, while the right instance is miss detected (false negative). Each prediction shows a case that exhibit one of the error types we categorize.}
    \label{fig:fp_visualization}
\end{figure}

Let $\mathcal{G}$ be the set of ground truth instances such that an instance $g^{(k)} = (g^{(k)}_l, g^{(k)}_t)$ consists of a label $g^{(k)}_l$ and temporal bounds $g^{(k)}_t$. Let $\mathcal{P}$ be the set of prediction segments such that a prediction $p^{(i)} = (p^{(i)}_s, p^{(i)}_l, p^{(i)}_t)$ consists of a score $p^{(i)}_s$, a label $p^{(i)}_l$, and a temporal extent $p^{(i)}_t$. A prediction $p^{(i)}$ is a \textbf{True Positive (TP)} if and only if $\exists g^{(k)} \in \mathcal{G}$ such that $p^{(i)}$ is the highest scoring prediction with $tIoU(g^{(k)}_t, p^{(i)}_t) \ge \alpha$ and $p^{(i)}_l = g^{(k)}_l$, where $\alpha$ is the tIoU threshold. Otherwise, the prediction is a \textbf{False Positive (FP)}. Suppose that $p^{(i)}$ is an FP prediction and $g^{(k)}$ is the ground truth instance with the highest tIoU with $p^{(i)}$.  We classify this FP prediction into five categories (see Figure \ref{fig:fp_visualization}).

\paragraph{Double Detection Error (DD).} A prediction that satisfies the tIoU threshold with a ground truth instance with the correct label, however, the ground truth instance is already matched with another prediction of a higher score. We identify this error due to the nature of the ActivityNet evaluation framework, which measures performance at high tIoU thresholds and penalizes double detections.
\begin{align}
    tIoU(g^{(k)}_t, p^{(i)}_t) \ge \alpha,\ g^{(k)}_l = p^{(i)}_l; \exists p^{(j)}\in \mathcal{P}, tIoU(g^{(k)}_t, p^{(j)}_t) \ge \alpha,\  p^{(j)}_s \ge p^{(i)}_s 
\end{align}

\paragraph{Wrong Label Error (WL).} A prediction that meets the tIoU threshold but incorrectly predicts the label of the ground truth instance. The source of this error is often a weakness in the action classification module.
\begin{align}
    tIoU(g^{(k)}_t, p^{(i)}_t) \ge \alpha \text{ and } g^{(k)}_l \ne p^{(i)}_l
\end{align}

\paragraph{Localization Error (LOC).} A prediction with the correct label that has a minimum $0.1$ tIoU and fails to meet the $\alpha$ tIoU threshold with the ground truth instance. The source of this error is typically a weakness in the localization module and/or the temporal feature representation.
\begin{align}
    0.1 \le tIoU(g^{(k)}_t, p^{(i)}_t) < \alpha \text{ and } g^{(k)}_l = p^{(i)}_l
\end{align}

\paragraph{Confusion Error (CON).} A prediction of the wrong label that has a minimum $0.1$ tIoU but does not meet the $\alpha$ tIoU threshold with the ground truth instance. This error is due to a combination of the same error sources in WL and LOC.
\begin{align}
    0.1 \le tIoU(g^{(k)}_t, p^{(i)}_t) < \alpha \text{ and } g^{(k)}_l \ne p^{(i)}_l
\end{align}

\paragraph{Background Error (BG).} A prediction that does not meet a minimum $0.1$ tIoU with any ground truth instance. This error could arise in large percentages due to a weakness in the prediction scoring scheme.
\begin{align}
    tIoU(g^{(k)}_t, p^{(i)}_t) < 0.1
\end{align}

\noindent Another error source of relevance to our analysis is the miss detection of ground truth instances, \ie~\textbf{False Negative (FN)}. In Section \ref{section:fn_analysis}, we analyze why some type of instances are typically miss detected by current algorithms.

%% file: sections/fp_analysis.tex
\section{False Positive Analysis}\label{section:fp_analysis}

In this section, we take the four state-of-the-art methods (SC, CES, IC, and BU) as an example to showcase our FP analysis procedure. 
First, we introduce the concept of a \textit{False Positive Profile}, the mechanism we employ to dissect a method's FP errors. 
Then, we present insights gathered from the methods' FP profiles. Finally, we investigate each error type's impact on the average-mAP$_N$.

\begin{figure}[t!]
    \centering
    \begin{subfigure}{0.99\linewidth}
        \includegraphics[width=1\linewidth]{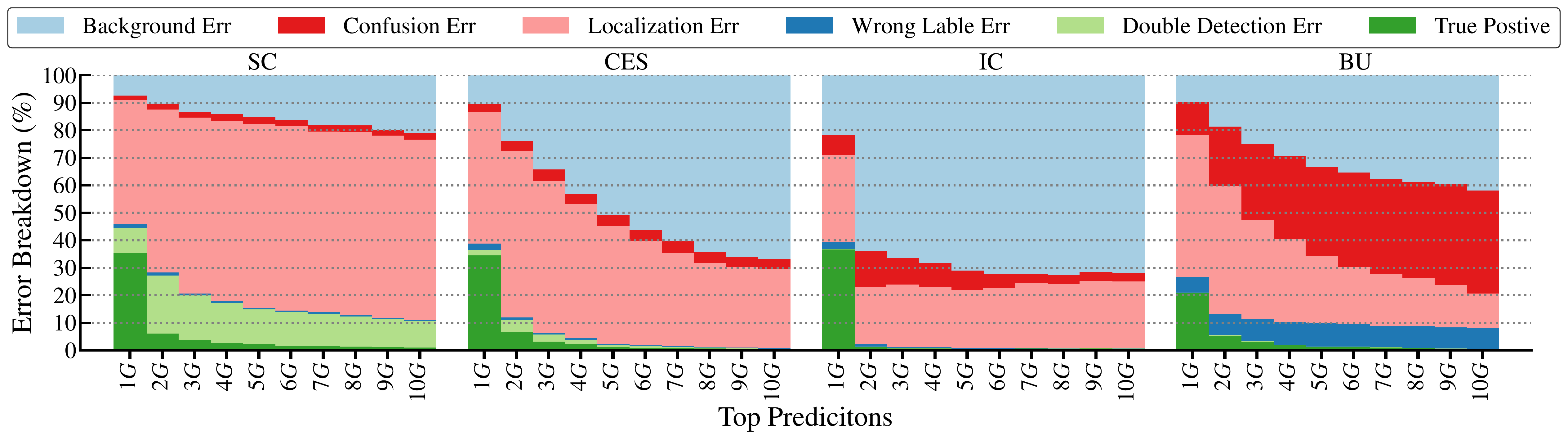}
    \end{subfigure}
    
    \begin{subfigure}{0.80\linewidth}
        \includegraphics[width=1\linewidth]{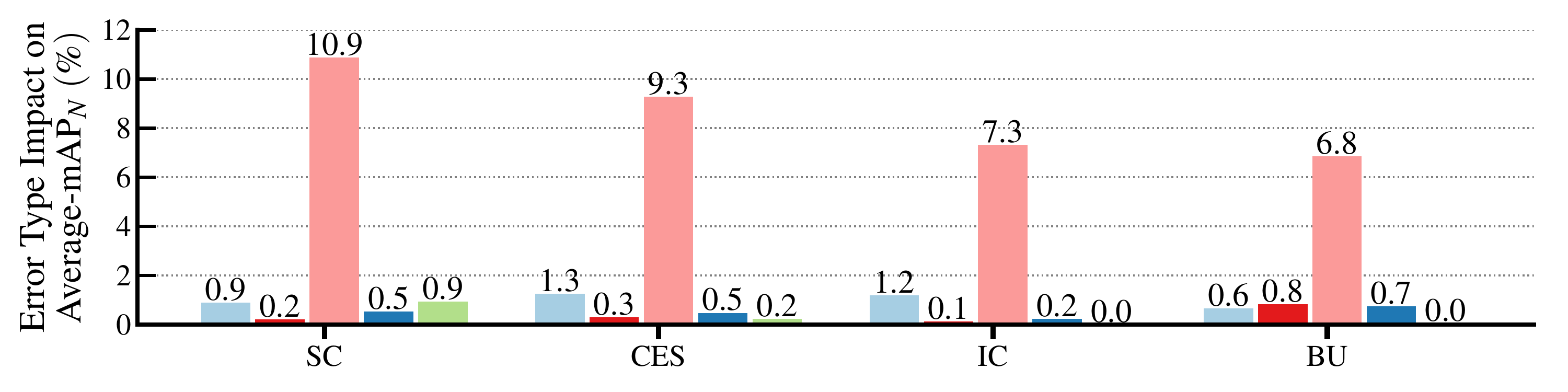}
    \end{subfigure}

    \caption{\textbf{Top:} The false positive profiles of the four methods. Each profile demonstrates the FP error breakdown in the top-$10G$ predictions. \textbf{Bottom:} The impact of error types on the average-mAP$_N$, \ie~the improvement gained from removing all predictions that cause each type of error. The \textit{Localization Error} (pink bar) has the most impact.
    }
    \label{fig:fp_profile_and_metric_improvement}
\end{figure}

\noindent\textbf{False Positive Profile.} The computation of average-mAP$_N$ relies inherently on the ranking of predictions. Thus, it is important to take the prediction score into account when analyzing FP errors. Thus, we execute our analysis on the error profile of the top-$10G$ predictions, where $G$ is the number of ground truth instances. We pick the top predictions in a per-class manner, \ie~we select the top-$10G_j$ predictions form class $j$, where $G_j$ is the number of instances in class $j$. Moreover, to see the trend of each error type, we split the top-$10G$ predictions into ten equal splits and investigate the breakdown of the five FP error types (defined in Section \ref{section:taxonomization}) in each split. The collection of these error breakdowns allow us to model the error rate of each type as a function of the prediction score. This intuitively allows us to examine the behaviors of the different detector components such as the classifier and scoring function.

We choose to focus on the top-$10G$ predictions instead of all predictions for the following four reasons: (i) $10G$ is sufficiently large to show trends in error types; (ii) Current state-of-the-art methods exhibit an extremely low normalized precision ($<0.05P_N$) beyond this large number of predictions. An error analysis at such low precision point is not insightful as predictions' quality degrades and the background errors dominate; (iii) The average-mAP$_N$ for the top-$10G$ is very close to the performance of all predictions (Table \ref{table:ANET_mAP_normalized}); and (iv) It is easier to compare the FP profiles of multiple methods on the same number of predictions.

\noindent\textbf{What can FP Profiles Tell us?} Figure \ref{fig:fp_profile_and_metric_improvement} (\textbf{Top}) shows the four methods' FP profiles. 
The top-$G$ predictions in all methods contain the majority of TP.
SC is the best in terms of background error rate, while IC is the worst since the majority of its predictions beyond the top-$G$ are background errors. This indicates a shortcoming in IC's scoring scheme. 
On the other hand, SC has a relatively high double detection error rate. We attribute this to the fact that SC is purely a proposal method (\ie~it is optimized for a high recall) combined with a video-level classifier that is independent of the proposal generation.
However, this double detection rate can be fixed by applying a stricter \textit{non-maximum-suppression} (NMS). 
Notably, errors due to incorrect labels (\ie~wrong label and confusion errors) are relatively small for the top three methods. This signals the strength of these methods' classifiers. At the same time, we can see a high wrong label and confusion errors for BU, indicating a weakness in BU's classifier.

\noindent\textbf{FP Categories Impact on the Average-mAP$_N$.} The insights we get from the FP profile help us to identify problems in algorithms, however, they do not tell us which problem we should prioritize fixing. In order to address this, we quantify the impact of an error type by measuring the average-mAP$_N$ after fixing that error, \ie~we calculate the metric after removing all predictions causing the given error type. Figure \ref{fig:fp_profile_and_metric_improvement} (\textbf{Bottom}) shows the impact of the five errors on the performance of the four methods. Fixing localization errors gives a significant boost to average-mAP$_N$, while fixing other error types provides limited improvements. This is a compelling evidence that localization error is the most significant error to tackle and that the research field should focus on addressing this error in order to advance detection algorithms.

%% file: sections/sensitivity.tex
\section{Average-mAP$_N$ Sensitivity}\label{section:sensitivity}

Typically, researchers design localization algorithms to tackle certain action characteristics. For example, multiple works have tried to capture the temporal context along the video \cite{DaiSZDC_iccv2017,GaoYSCN_iccv2017,LaptevMSR_cvpr08,sigurdsson2017asynchronous,ZhaoXWWTL_iccv2017} as a proxy for the localization of the actions. Indeed, the recent SSN architecture, presented in this study as CES, is the latest successful work in this school of thought.
In this case, the architecture not only describes each segment proposal, as typically done by template based methods, but it also represents the adjacent segments around to influence the instances localization.
Although, these ideas are well motivated, it is unclear if changes in performance with respect to a single metric actually corresponds to representative changes on instances with the characteristics of interest. In that sense, another important component in the diagnosis of localization algorithms is the analysis of AP variation with respect to the actions characteristics.

\begin{figure}[t!]
    \centering
    \begin{subfigure}{0.75\linewidth}
        \includegraphics[width=1\linewidth]{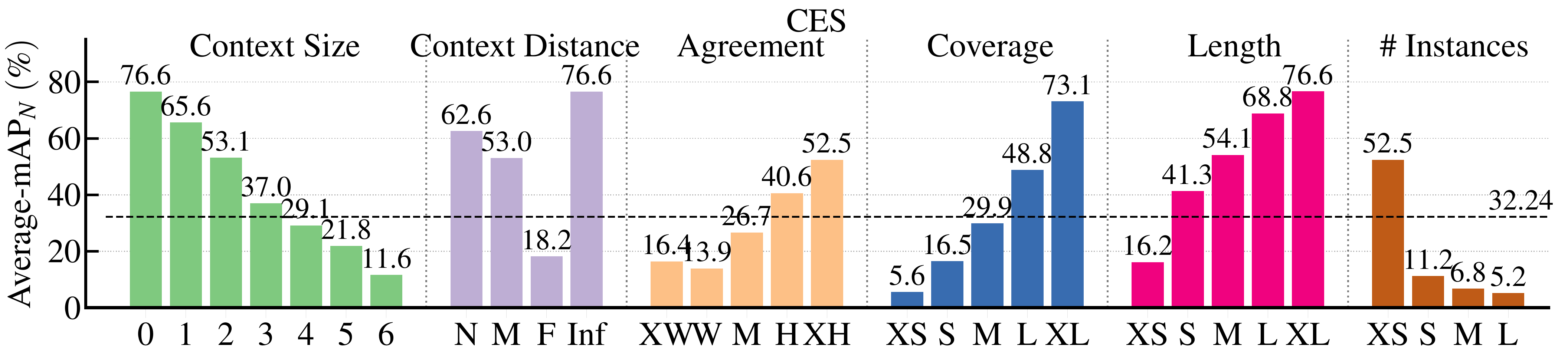}
    \end{subfigure}
    \begin{subfigure}{0.24\linewidth}
        \includegraphics[width=1\linewidth]{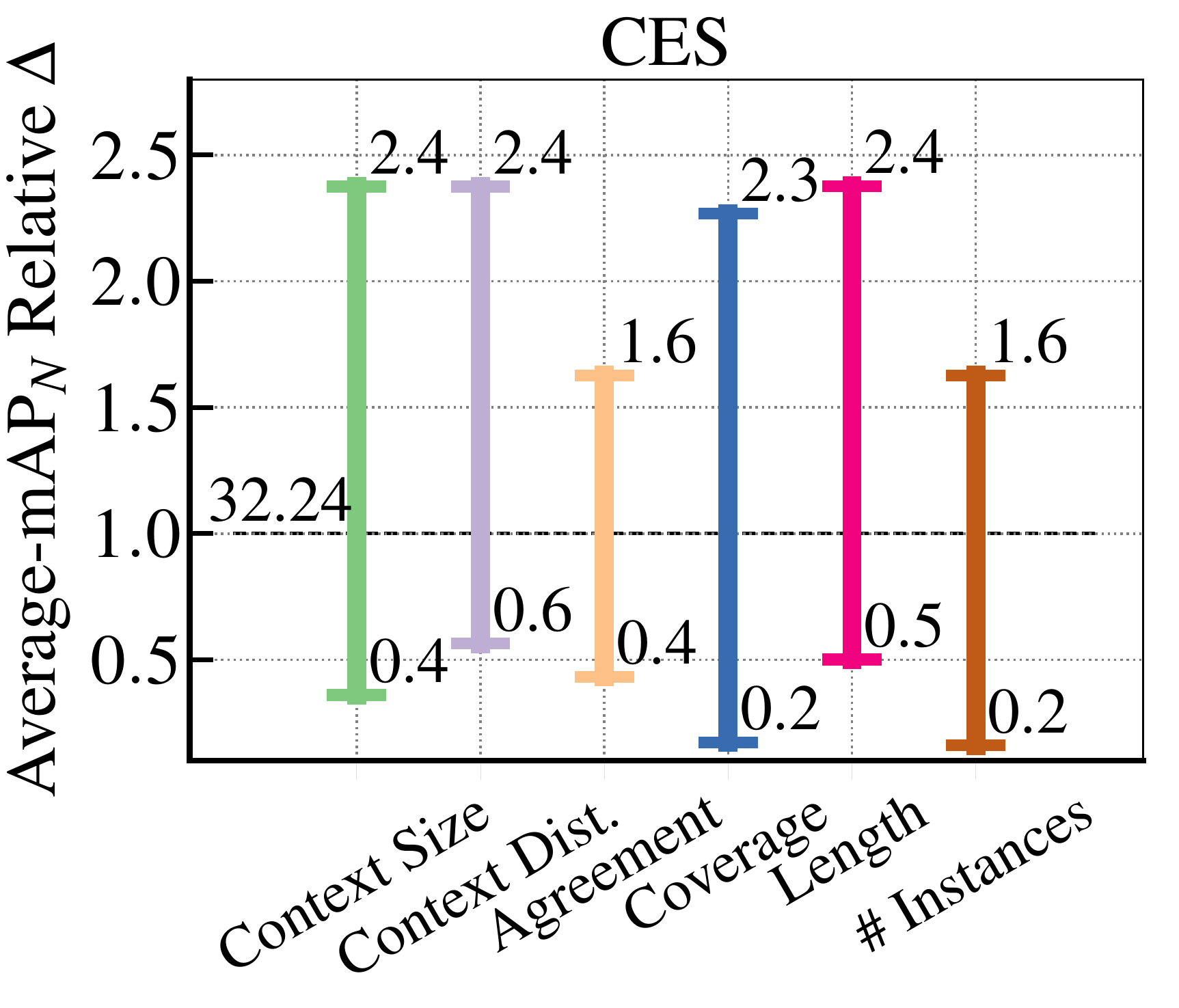}
    \end{subfigure}
    \caption{
        \textbf{Left:} The detailed sensitivity CES's average-mAP$_N$ to action characteristics. The dashed line is the overall performance. Each bar measures the average-mAP$_N$ on a subset of ActivityNet for which a particular action characteristic holds. \textbf{Right:} The sensitivity profile summarizing the left figure. The difference between the max and min average-mAP$_N$ represents the sensitivity, while the difference between the max and the overall average-mAP$_N$ denotes the impact of the characteristic.
    }
    \label{fig:sensitivity}
\end{figure}

Figure \ref{fig:sensitivity} (\textbf{Left}) shows CES's performance variations over all the action characteristics described in Section \ref{section:dataset_characterization}. Each bar represents the performance after dropping all the instances that do not exhibit a particular characteristic, and the dash bar represents the performance of the method over all the instances in the dataset. In contrast with the analysis of FP profiles (Section \ref{section:fp_analysis}), all four methods exhibit similar variation trends across multiple action characteristics. Refer to the \textit{supplementary material} for the other methods' sensitivity figures.

Among the interesting patterns to highlight, we find instances where humans tend to agree more on the starting and end of the action translates to gains in performance (H-XH agreement in the figure), while the opposite behavior shows a drop in performance. 
This correlation is a bit surprising considering that the models are not trained with multiple annotations per instance.
Unfortunately, we do not find any concluding evidence to explain this interesting correlation besides the nature of the instances or the bias of the dataset.
Similarly, instances where an action occurs naturally surrounded by enough temporal evidence that reinforces its presence are associated with drops in performance (context size of 5-6 in the figure).
We argue that this is due to the presence of similar actions around the instances which creates a confusion and impedes precise boundaries positioning around the instance.
In terms of coverage and instance length, the results are intuitive and easy to interpret. Short instances, either absolute in time or in relationship to the video length, tend to be more difficult to detect. This could be a consequence of the coarse temporal structure used by the algorithms to accumulate temporal evidence across the video.

Figure \ref{fig:sensitivity} (\textbf{Right}) summarizes, in a sensitivity profile, the variations in CES's average-mAP$_{N}$ for each characteristic group, as well as the potential impact of improving the robustness. 
According to our study, all the methods exhibit a similar trend, they are more sensitive to variations in the temporal context, coverage, and length compared to variations in the agreement and number of instances.
Based on experiments with ideal classifiers, \cite{SigurdssonRG17} hypothesizes the temporal agreement between annotators is not a major roadblock for action localization.
Interestingly, our diagnostic analysis shows the first experimental evidence corroborating this hypothesis.
Considering the small positive and negative impacts of these instances, efforts to improve in this area must be validated carefully such that improvements do not come from more common and easier cases.
Our analysis also justifies researchers' focus on designing algorithms that exploit temporal context for localization. Three out of the four models studied here would benefit the most by improving the contextual reasoning along the temporal scale.

%% file: sections/fn_analysis.tex
\section{False Negative Analysis}\label{section:fn_analysis}

\begin{figure}[t!]
    \centering
    
    \begin{subfigure}{1\linewidth}
        \includegraphics[width=0.99\textwidth]{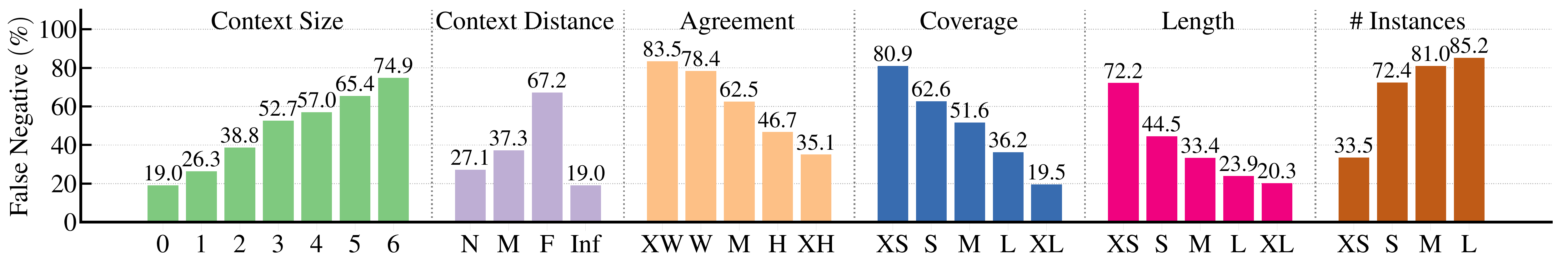}
    \end{subfigure}
    
    \begin{subfigure}{0.21\linewidth}
        \includegraphics[width=1\linewidth]{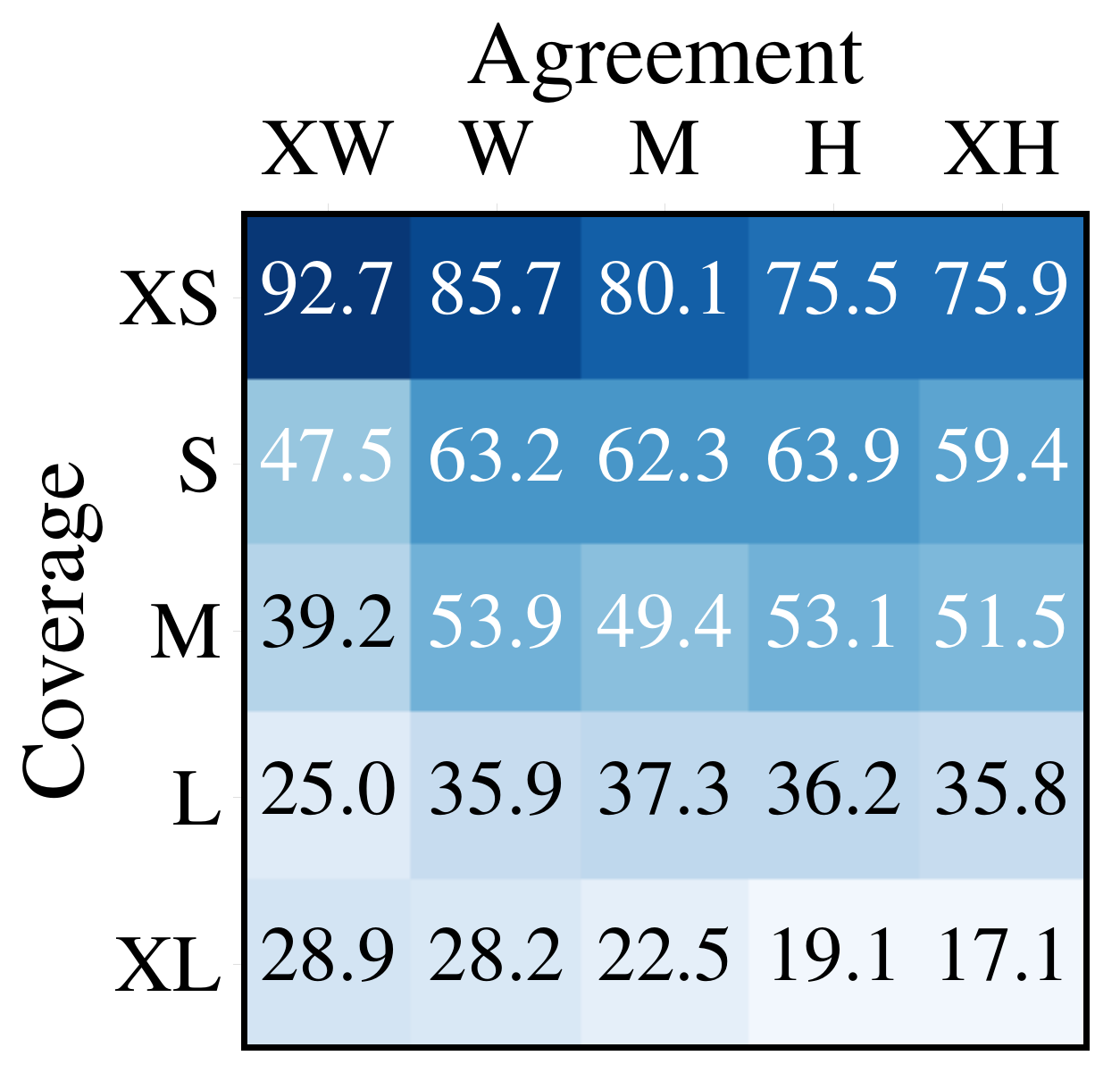}
    \end{subfigure}
    \begin{subfigure}{0.36\linewidth}
        \includegraphics[width=1\linewidth]{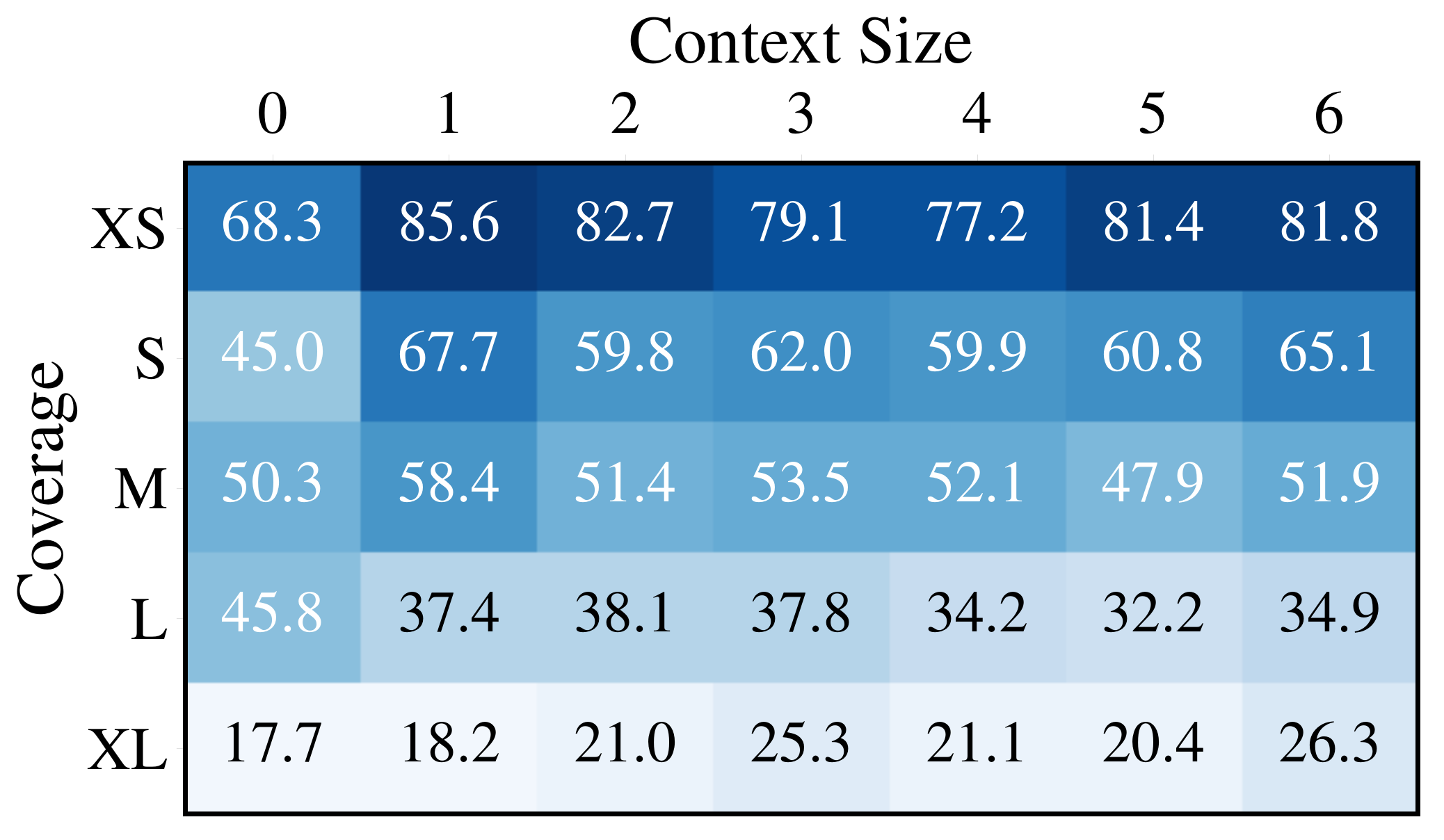}
    \end{subfigure}
    \begin{subfigure}{0.36\linewidth}
        \includegraphics[width=1\linewidth]{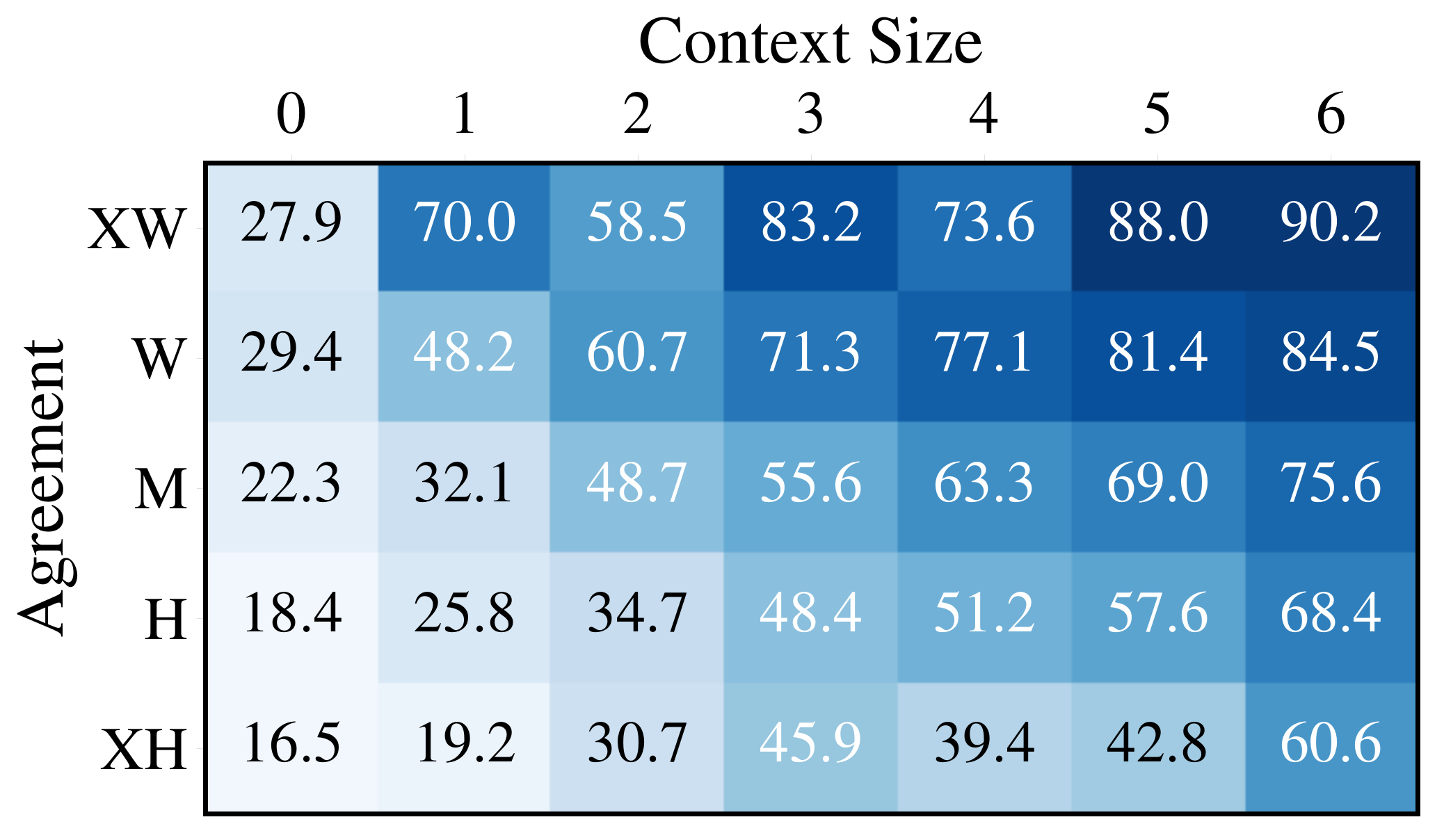}
    \end{subfigure}

    \caption{Average false negative rate across algorithms for each characteristic \textbf{(Top)} and three pairs of characteristics \textbf{(Bottom)}. We observe characteristics such as XS coverage and XW agreement are hard to detect individually and when paired with others. Differently, instances with XL coverage and XH agreement are relatively easy.}
    \label{fig:pairwise_distribution}
\end{figure}

So far, we have only considered the types of FP errors introduced by the detections algorithms, and the characteristics of the actions that introduce more variations in the performance. On the other hand, it is insightful to study what makes an action instance difficult to detect, even at minuscule confidence threshold. Towards this end, we compute the percentage of missed detections instances for each algorithm and group them according to the action characteristics defined in Section \ref{section:dataset_characterization}. For this purpose, we consider that an action instance is miss detected if we do not find a matching detection at a precision level higher than $0.05 P_{N}$. Figure \ref{fig:pairwise_distribution} (\textbf{Top}) summarizes our findings. In the interest of saving space, we average the results across multiple algorithms (refer to the \textit{supplementary material} for the results of each algorithm by itself). The first observation that we can grasp from the results is its inverse relationship with the sensitivity profiles shown in Figure \ref{fig:sensitivity}. For example, the drops in performance we observed for instances with extremely weak agreement, low coverage, short length, or high temporal context size match the evidence that the algorithms struggle to retrieve such instances.
On the other hand, we can appreciate that algorithms are struggling to find multiple instances per videos. Note how the amount of missed detections increase more than double due to the presence of another instance in the video. This is definitely an area where methods should focus on to mitigate the negative impact in performance. 
For context distance, the pattern is intuitive given that the increase in cumulative size of the \textit{context glimpses} correlates with the spread of context and confusion in time. Thus, the chance of delimiting the start and end of the instance get worse.

Finally, we also find some interesting patterns in the FN rate at the intersection between two groups of characteristics. Figure \ref{fig:pairwise_distribution} (\textbf{Bottom}) compactly summarizes those in a similar fashion.
It is interesting how particular pairwise combinations such as low coverage (XS) - large context size (6), extremely weak agreement (XW) - large context size (6), and low coverage (XS) - extremely weak agreement (XW) are very difficult to detect, even when some are well represented in the dataset.
Similarly, we find pairs involving high agreement (XH), small context (0), and high coverage (XL) relatively easy to detect.
Finally, we find some interesting contours in the pairwise interactions, \eg~the percentage of FN diffuses in the matrix of agreement v.s. context size as we move from the top right corner to the bottom left in a non-smooth way.

%% file: sections/discussion.tex
\section{Discussion and Conclusion}

We introduced a novel diagnostic tool for temporal action localization and demoed its application by analyzing four approaches in the latest ActivityNet action localization challenge.
We showed how our proposed methodology helps detection methods not only to identify their primary sources of FP errors but also to reason about their miss detections. 
We provided a detailed categorization of FP errors, which is tailored for action localization specifically. Using this categorization, we later defined our proposed \textit{False Positive Profile} analysis. We found that the FP profile varies across methods. Some of the techniques exhibited shortcomings in their scoring function, while others showed weaknesses in their action classifier. We also investigated the impact of each error type, finding that \textit{all} detectors are strongly hurt by localization errors.
We conducted an extensive dataset characterization, which empowered us with a deeper understanding of what makes an action instance harder to localize. We introduced and collected six new action characteristics for the ActivityNet dataset, namely, context size, context distance, agreement, coverage, length, and the number of instances. We measured methods' sensitivities to these action characteristics. We observed all methods are very sensitive to temporal context. Also, we showed that temporal agreement between annotators is not a significant barrier towards improving action localization.
For future work, we plan to explore new metrics for action localization that incorporate the inherent ambiguity of temporal action boundaries. In our \textit{supplementary material}, we present a preliminary study that exploits the newly collected temporal annotations to ease the strict performance computation current evaluation frameworks use.
With the release of our diagnostic tool, we aim to empower the temporal action localization community with a more in-depth understanding of their methods' failure modes. Most importantly, we hope that our work inspires the development of innovative models that tackle the current flaws of contemporary action localization approaches.\\

\noindent\textbf{Acknowledgments.}
This publication is based upon work supported by the King Abdullah University of Science and Technology (KAUST) Office of Sponsored Research (OSR) under Award No. OSR-CRG2017-3405.